\documentclass[electronics,article,accept,pdftex,moreauthors]{Definitions/mdpi} 

\firstpage{1} 
\pubvolume{1}
\issuenum{1}
\articlenumber{0}
\pubyear{2025}
\copyrightyear{2025}
\externaleditor{Giuseppe Prencipe} 
\datereceived{23 February 2025} 
\daterevised{3 April 2025} 
\dateaccepted{10 April 2025 } 
\datepublished{ } 
\hreflink{https://\linebreak doi.org/} 

\Title{Knitting Robots: A Deep Learning Approach for Reverse-Engineering Fabric Patterns} 
\TitleCitation{Knitting Robots: A Deep Learning Approach for Reverse-Engineering Fabric Patterns}

\Author{{Haoliang Sheng} $^{1}$, Songpu Cai $^{1}$, Xingyu Zheng $^{1,2}$ and {Mengcheng} Lau $^{1,}$*$^{}$\orcidA{}}

\AuthorNames{Haoliang Sheng, Songpu Cai, Xingyu Zheng and Mengcheng Lau}

\isAPAStyle{%
       \AuthorCitation{Sheng, H., Cai, S., Zheng, X., \& Lau, M.}
         }{%
        \isChicagoStyle{%
        \AuthorCitation{Haoliang Sheng, Songpu Cai, Xingyu Zheng, and Mengcheng Lau.}
        }{
        \AuthorCitation{Sheng, H.; Cai, S.; Zheng, X.; Lau, M.}
        }
}

\address{%
$^{1}$ \quad Bharti School of Engineering and Computer Science, Laurentian University, Sudbury, ON {P3E 2C6}, Canada; hsheng@laurentian.ca (H.S.); scai1@laurentian.ca (S.C.); xzheng2@laurentian.ca (X.Z.)\\
$^{2}$ \quad Shanghai International Fashion Education Centre, {Shanghai} 200060, China}

\corres{Correspondence: mclau@laurentian.ca; Tel.: +1-705-675-1151}

\abstract{Knitting, a cornerstone of textile manufacturing, is uniquely challenging to automate, 
	particularly in terms of converting fabric designs into precise, machine-readable instructions. This research bridges the gap between textile production and robotic automation by proposing a novel deep learning-based pipeline for reverse knitting to integrate vision-based robotic systems into textile manufacturing. The pipeline employs a two-stage architecture, enabling robots to first identify {\textit{front labels}} before inferring {\textit{complete labels}}, ensuring accurate, scalable pattern generation. By incorporating diverse yarn structures, including single-yarn (\textit{sj}) and multi-yarn (\textit{mj}) patterns, this study demonstrates how our system can adapt to varying material complexities. Critical challenges in robotic textile manipulation, such as label imbalance, underrepresented stitch types, and the need for fine-grained control, are addressed by leveraging specialized deep-learning architectures. This work establishes a foundation for fully automated robotic knitting systems, enabling customizable, flexible production processes that integrate perception, planning, and actuation, thereby advancing textile manufacturing through intelligent robotic automation.}

\begin{document}


\section{Introduction}

\begin{figure}[H]
\includegraphics[width=7 cm]{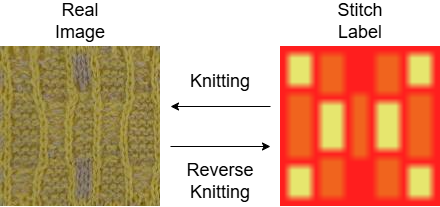}
\caption{Reverse knitting.\label{FIG: reverse knitting}}
\end{figure}

Knitting has long been a cornerstone of textile manufacturing, from traditional hand-crafted designs to industrial whole-garment knitting machines. In conventional processes, a predefined \textit{{stitch label}} serves as the input for knitting machines like those controlled by the \textit{m1plus} \cite{bohm2022knitting} {programming} language. These inputs generate corresponding fabrics (outputs), which can be visualized or verified through images of the resulting physical objects. However, the process of creating knittable programs or \textit{stitch labels} from existing images of knitted patterns remains a complex challenge. This gap defines the scope of \textit{{reverse knitting}}, as shown in Figure~\ref{FIG: reverse knitting}, which is an emerging paradigm that utilizes real images of knitted fabrics as its input and outputs knittable stitch labels through deep learning models. Such systems hold immense potential to transform the design and manufacturing pipeline, especially in terms of customization and replication tasks.

The primary objective of this project is to develop a two-stage pipeline for \textit{reverse knitting}, leveraging deep learning to transform real images of knitted fabrics into knittable stitch labels. This pipeline includes a \textit{generation phase} for predicting \textit{front labels} and an \textit{inference phase} for generating \textit{complete labels}. This research aims to improve the accuracy of label prediction for diverse yarn structures, addressing the challenges of single-yarn (\textit{sj}) and multi-yarn (\textit{mj}) knitting patterns through specialized architectures. Additionally, this study evaluates the pipeline's performance across realistic scenarios, focusing on its ability to handle imbalanced datasets and generalize effectively. This research explores three key questions. The first focuses on pipeline design, investigating how a two-stage deep learning pipeline can be effectively structured to transform real fabric images into knittable stitch labels. The second addresses performance optimization, examining the impact of design choices such as stricter loss functions, yarn-specific training, and different model architectures on the accuracy and robustness of \textit{front label} and \textit{complete label} predictions. Lastly, this research evaluates the pipeline’s ability to handle realistic constraints, such as imbalanced datasets and unknown yarn types, while maintaining high performance across various application scenarios.

Early efforts, such as those of Kaspar et al. \cite{kaspar2019neural}, introduced the Refiner+Img2prog architecture using a SimGAN framework \cite{shrivastava2017learning} to refine\textit{ real images} into synthetic \textit{rendering images} for \textit{front label} prediction. However, challenges like dataset imbalance (e.g., FK dominance) and the “one-to-many” problem—where identical instructions yield varied visuals—limit precision and are often exacerbated by multiple-instance learning (MIL) strategies \cite{kaspar2019neural}. Stricter loss functions have since been explored to enhance accuracy. Trunz et al. \cite{trunz2024} and Melnyk \cite{melnyk2022} extended GAN-based approaches \cite{goodfellow2014generative, mirza2014conditional, mao2017least} to model yarn properties and fabric styles, while CNNs \cite{leCun1989} excel at capturing spatial stitch relationships and LSTMs \cite{hochreiter1997} address row \mbox{dependencies \cite{scheidt2020}}. However, these works focus more on creatively generating stitches, unlike Kaspar’s Inverse Knitting, which only translates fabric images into \textit{stitch labels}.

For the \textit{inference phase}, inter-stitch logic is critical. Yuksel et al. \cite{yuksel2012stitch} pioneered stitch meshes, which were later refined by Wu et al. \cite{wu2018a, wu2018b} for knittability. Conditional Random Fields (CRFs) \cite{lafferty2001} model probabilistic dependencies, though long-range limitations persist \cite{he2016deep}. Residual architectures \cite{he2016deep} and 3D-to-instruction methods \cite{narayanan2018} further enhance \textit{complete label} generation. Yet, gaps remain: imbalanced datasets underrepresent rare labels (e.g., “E”, “V”) \cite{kaspar2019neural, trunz2024, melnyk2022}, multi-yarn scalability is limited \cite{yuksel2012stitch, wu2018a}, and computational efficiency lags are common with resource-heavy models \cite{hochreiter1997, lafferty2001, scheidt2020}.

Unlike Transformers, which favor global features but demand a large amount of resources, our residual CNNs, UNets, and multi-layer CNNs prioritize local dependencies, as \textit{stitch labels} must work with their neighbors \cite{yuksel2012stitch}, using 3 $\times$ 3 kernels effectively. GANs \cite{kaspar2019neural} remain cost-efficient, while GNNs, though relational, face convergence issues in dense grids and \mbox{CNNs favor stability}.

The rest of this paper is organized as follows: Section \ref{sec2} details the data collection and preparation process, including the acquisition of ground truth data and the categorization of yarn and stitch types. Section \ref{sec3} presents the proposed two-stage deep learning architecture, describing the \textit{generation phase} for \textit{front label} prediction and the \textit{inference phase} for \textit{complete label }generation. Section \ref{sec4} describes the experimental setup, including the four key usage scenarios, and evaluates the model's performance across different yarn types and stitch patterns. Section \ref{sec5} discusses the results, highlighting the model's strengths and limitations, while Section \ref{sec6} concludes the study and outlines future research directions. Finally, Appendices \ref{chap:usage_scenarios} and \ref{chap:environment_configuration} provide additional details about the model's implementation and the experimental environment setup.

\section{Data Collection and Preparation}\label{sec2}

\begin{figure}[H]
\includegraphics[width=6 cm]{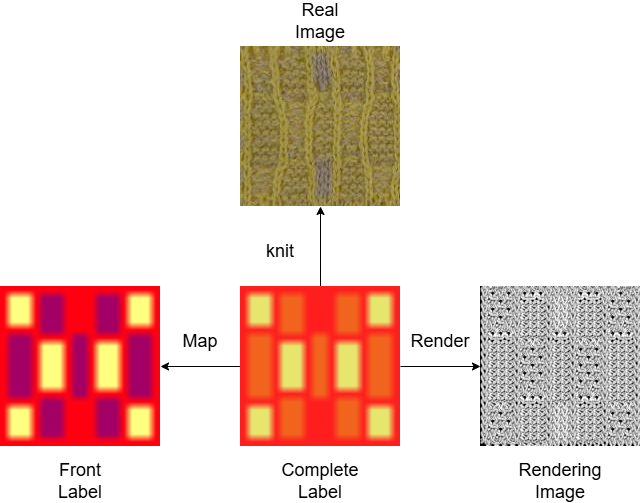}
\caption{Data processing workflow.\label{FIG: data process}}
\end{figure}

The data collection and preparation process involves acquiring ground truth data and defining yarn and stitch concepts. As illustrated in Figure~\ref{FIG: data process}, the ground truth data consist of four key components: the \textit{complete label}, designed by textile experts as the foundation for other data types; the \textit{front label}, which is derived from the \textit{complete label} through mapping; the \textit{rendering image}, generated by rendering the \textit{complete label}; and the real image, produced by knitting the \textit{complete label} into physical fabric. These data types are systematically derived through processes such as mapping the \textit{complete label} to the \textit{front label}, rendering these into images, and knitting these into physical fabrics.

Our dataset is categorized by yarn types and stitch patterns. 
The yarn types include single-yarn samples (\textit{sj}) and multi-yarn samples (\textit{mj}), which are further divided into 2\emph{j}, 3\emph{j}, and 4\emph{j} for fabrics containing two, three, and four yarn colors, respectively. However, practical limitations restrict structures to less than four colors. 
Each dataset sample consists of a 20 $\times$ 20 grid of stitches labeled with stitch types like \textit{FK}, \textit{BK}, and others, which represent the structural details of the fabric.

\begin{figure}[H]
\includegraphics[width=7 cm]{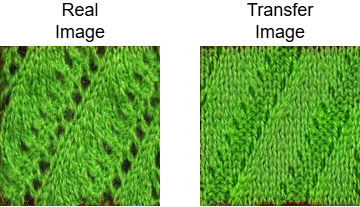}
\caption{Comparison of transfer image with real image.\label{FIG: transfer image}}
\end{figure}

To address the limited availability of \textit{real images} in the Kaspar dataset, which contains only 1043 front-facing fabric images, data augmentation is performed using transfer images (Figure~\ref{FIG: transfer image}). Both rendering images and transfer images are generated from the \textit{complete label} using the \textit{m1plus} software provided by Stoll \cite{bohm2022knitting}. These \textit{transfer images} serve as substitutes for \textit{real images} captured through physical knitting and photography, allowing for a larger and more balanced dataset. This augmentation strategy significantly enhances the dataset's consistency and availability, ensuring the more reliable training and evaluation of the models.

The \textit{complete label }is not provided by the designers as raw data but is derived from the \textit{colored complete label }through label aggregation. It incorporates both structure and color tags. For example, in multi-yarn (\textit{mj}) cases containing two yarns (\textit{2j}), labels such as \textit{(1/2j)FK,MAK} and \textit{(2/2j)FK,MAK} are aggregated into \textit{FK,MAK}. Here, \textit{1/2j} indicates the first yarn in a two-yarn structure, while \textit{2j} specifies the yarn type. This label aggregation simplifies structure recognition by focusing on \textit{stitch labels} without considering color-specific information. Notably, color recognition is outside the scope of this project, which centers on identifying the stitch structures required for knitting instructions.

\subsection{Front Label Acquisition}
\begin{figure}[H]
\subfloat[\centering]{\includegraphics[width=6cm]{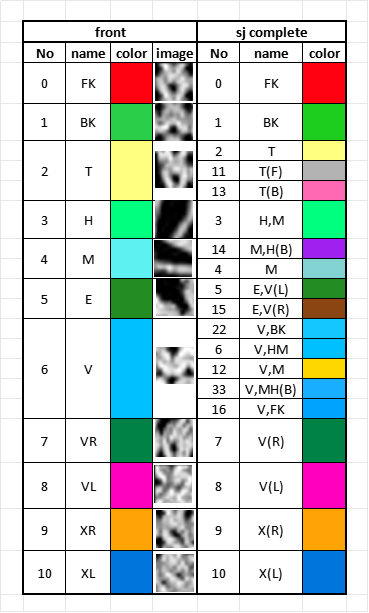}\label{fig:maps sub1}}
\hspace{20pt}
\subfloat[\centering]{\includegraphics[width=6cm]{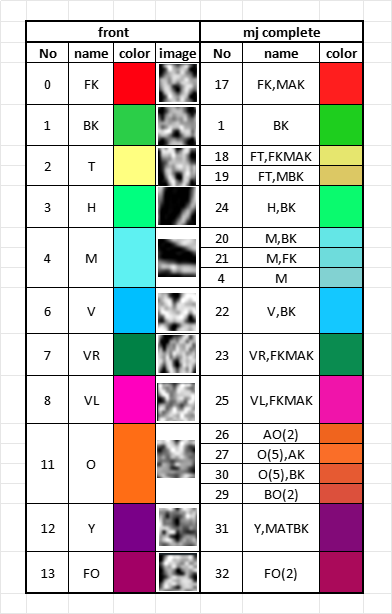}\label{fig:maps sub2}}
\caption{{Mapping} from front label to complete label. The "No" column represents the numerical identifiers assigned to each stitch type. The "name" column lists the abbreviated names of the stitch types. The "color" column indicates the encoded colors associated with each stitch type, which are used for visualization purposes. The "image" column (left table) is reserved for displaying physical representations or diagrams of the corresponding stitch types. (\textbf{a}) \textit{sj} map; (\textbf{b}) \textit{mj} map.\label{FIG: maps}}
\end{figure}

The \textit{front label} is obtained using predefined mappings, as illustrated in Figure~\ref{FIG: maps}. These contain the label number (numerical values used for training), label name (technical stitch types such as \textit{FK} and \textit{BK}), and label color and are employed for intuitive visualization of the 20 $\times$ 20 stitch grid during testing. 
Additionally, the front label provides a representation within the \textit{rendering image}, ensuring easy identification. In contrast, the\textit{ complete label }focuses on generating knittable instructions and includes the label number, label name, and label color, emphasizing the structural information required for producing feasible patterns.

A key distinction exists between the\textit{ front label} and the \textit{complete label}. The \textit{front label} simplifies the recognition of visual patterns in \textit{real images }or \textit{rendering images}, enabling a model to map directly from the visual domain. Meanwhile, the \textit{complete label} requires an inference step to generate knittable instructions, making it more suited to logical reasoning tasks rather than visual recognition. To ensure effective model training, separate mappings are provided for single-yarn (\textit{sj}) and multi-yarn (\textit{mj}) structures, as shown in Figure~\ref{FIG: maps}a,b. However, because the \textit{front label} exhibits minimal structural differences between \textit{sj} and \textit{mj} cases, combined training is feasible when using front labels.

\subsection{Rendering Image Acquisition}
\textit{Rendering images} were synthesized using the \textit{m1plus} software by Stoll \cite{bohm2022knitting}, which simulates virtual fabric structures using \textit{complete labels}. The process involves inputting the\textit{ complete label} into the \textit{m1plus} software, which generates a corresponding \textit{rendering image}. These images serve as intermediary outputs for downstream tasks, bridging the gap between virtual designs and real-world knitting data.

\subsection{Real Image Acquisition}
\begin{figure}[H]
\includegraphics[width=7 cm]{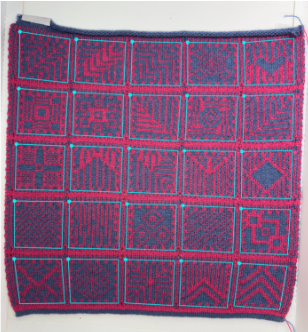}
\caption{Physical sheet.
	\label{FIG: raw real}}
\end{figure}

\textit{Real images} were generated using Stoll's computerized flat knitting machines, which are programmed with outputs from the \textit{m1plus} software. To improve efficiency, multiple samples were knitted together on a single sheet, as illustrated in Figure~\ref{FIG: raw real}. The Region of Interest (ROI) for each sample was then manually marked to distinguish between the different patterns within the sheet. Finally, physical images were captured using an iPhone 15 Pro Max in RAWMAX photo mode within a controlled lightbox environment, ensuring high-quality imaging for further processing and analysis.

\unskip

\subsection{Distribution of Yarn Types and Stitches}
\begin{figure}[H]
\includegraphics[width=7 cm]{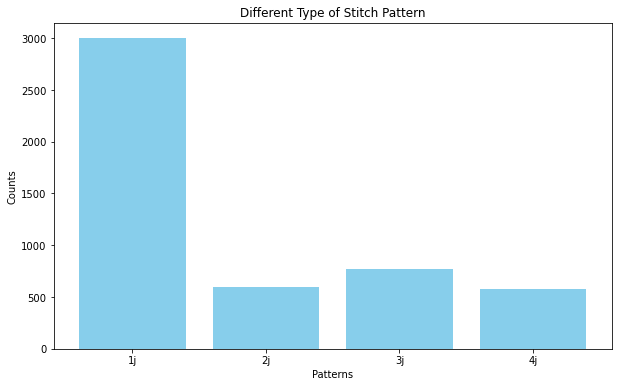}
\caption{Yarn type distribution.\label{FIG: yarn dist}}
\end{figure}

The \textit{distribution of yarn types}, illustrated in Figure~\ref{FIG: yarn dist}, includes a total of 4950 samples. The dataset consists of 3000 single-yarn (\textit{sj}) samples, evenly distributed across 10 subcategories, Hem, Move1, Miss, Cable1, Links2, Move2, Cable2, Mesh, Tuck, and Links1, with approximately 300 samples in each category. Additionally, there are 1950 multi-yarn (\textit{mj}) samples. Although a larger dataset of 12,392 \textit{sj} samples from Kaspar et al. \cite{kaspar2019neural} was available, the decision to limit the \textit{sj} samples to 3000 was made to maintain a balance with the smaller number of \textit{mj} samples. This curated subset ensures a representative distribution of both categories for effective model training.

\begin{figure}[H]
\centering
\begin{adjustwidth}{-\extralength}{0cm}
\subfloat[\centering]{\includegraphics[width=8.8cm]{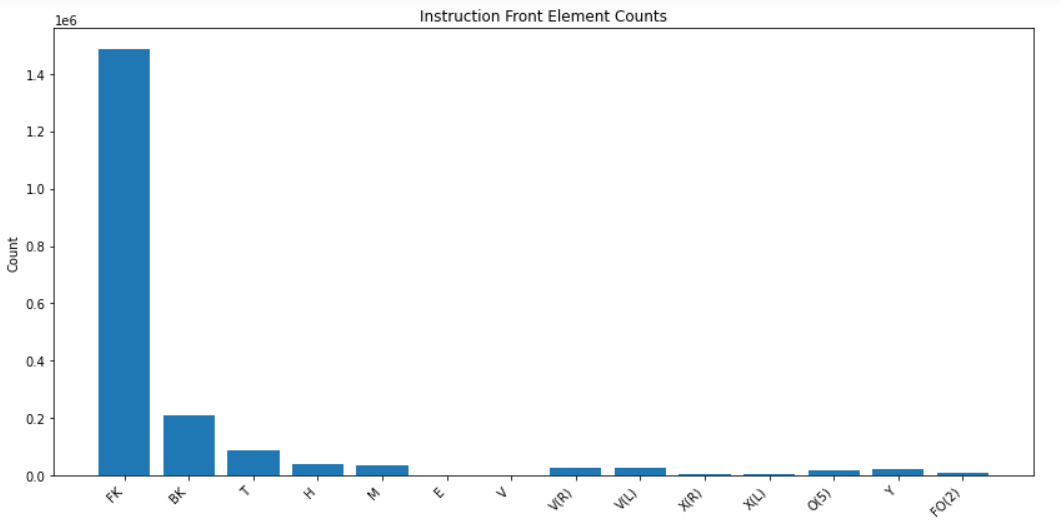}}
\hspace{20pt}
\subfloat[\centering]{\includegraphics[width=8.8cm]{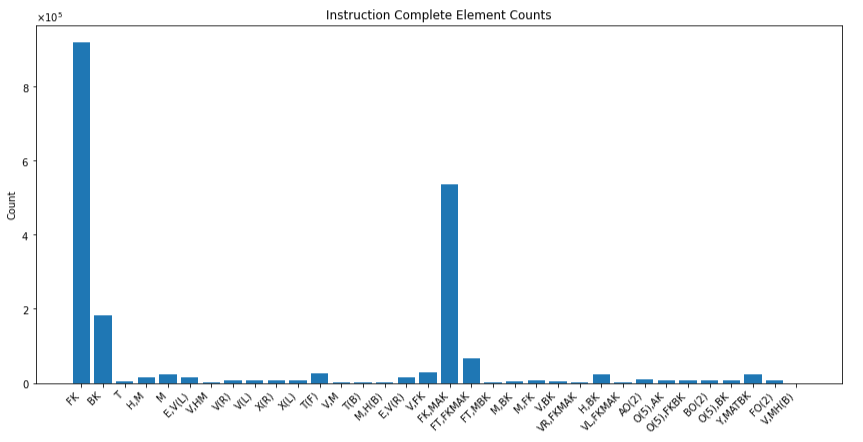}}
\end{adjustwidth}
\caption{{Stitch} distribution for (\textbf{a}) front labels and (\textbf{b}) complete labels.
	\label{FIG: stitch dist}}
\end{figure}

The \textit{stitch distribution} of the dataset reveals a significant label imbalance, as depicted in Figure~\ref{FIG: stitch dist}. For the \textit{front labels}, the \textit{FK} label dominates the dataset, accounting for approximately 75\% of the samples, while other labels remain underrepresented. Similarly, for the \textit{complete labels}, \textit{FK} makes up about 46.5\% of the samples, followed by \textit{FK}, with \textit{MAK} at approximately 27.1\% and the other labels being comparatively rare. This imbalance reflects the real-world variability in textile patterns and highlights the need for robust training strategies to mitigate its impact on \mbox{model performance}.
\vspace{-6pt}

\section{Model Architecture}\label{sec3}
The architecture we propose for generating knitting instructions from fabric images is structured to address the key limitations identified in previous methods, particularly in the approach used by Kaspar et al. (2019) \cite{kaspar2019neural}. In contrast to Kaspar et al., who utilized a single \textit{generation phase} to directly predict 17 single-yarn (\textit{sj}) labels, our methodology introduces an additional \textit{inference phase} that significantly enhances the scope and precision of the generated \mbox{knitting instructions}.

In our approach, the \textit{generation phase} initially predicts 14 distinct \textit{front labels} instead of final \textit{sj }labels. These \textit{front labels} are carefully chosen to capture the visual differences that are more discernible in the fabric images, making them easier for the model to identify accurately. By focusing on these foundational features, we simplify the initial prediction task, which improves the robustness and accuracy of \textit{front label} generation.

The \textit{inference phase} then takes these 14 \textit{front labels} and processes them through a \textit{residual model} to generate a comprehensive set of 34 \textit{complete labels}. This \textit{complete label} set is not only compatible with single-yarn knitting (\textit{sj}) but is also designed to support multi-yarn (\textit{mj}) knitting applications, enabling the production of more intricate and multi-colored patterns. This two-step approach—starting with visually distinctive front labels and refining them into complete knittable labels—represents a major advancement in automated knitting instruction generation. This architecture thus enhances both the scalability and flexibility of the model, making it suitable for a broader range of textile applications.

\begin{figure}[H]
\subfloat[\centering]{\includegraphics[width=7.7cm]{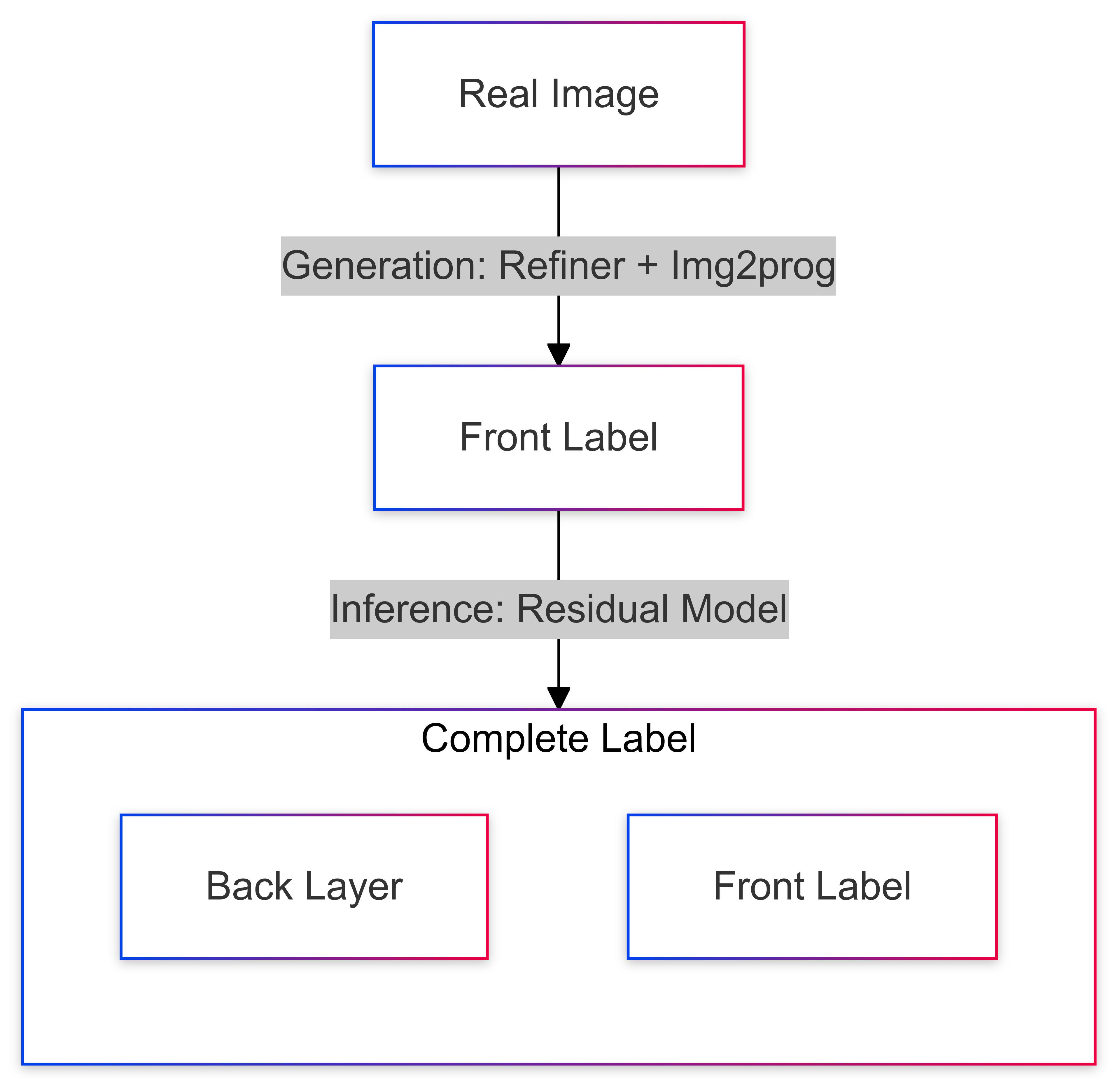}}
\hspace{20pt}
\subfloat[\centering]{\includegraphics[width=4.4cm]{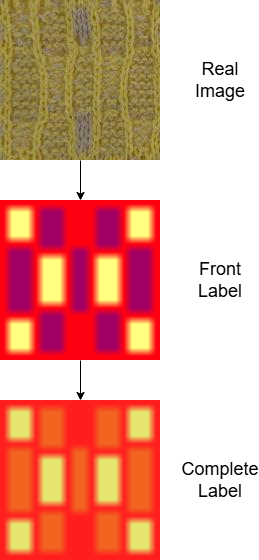}}
\caption{Overall architecture: (\textbf{a}) architecture diagram; (\textbf{b}) input and output.\label{FIG: overall architecture}}
\end{figure}

The overall architecture can be visualized in the  schematic provided below (Figure~\ref{FIG: overall architecture}), which illustrates the flow from \textit{real image} to \textit{front label} and finally to \textit{complete label} through the proposed \textit{generation and inference phases}.
\vspace{-9pt}

\subsection{Front Label Generation: Refiner and Img2prog}
\begin{figure}[H]
\begin{adjustwidth}{-\extralength}{0cm}
\subfloat[\centering]{\includegraphics[scale=0.1]{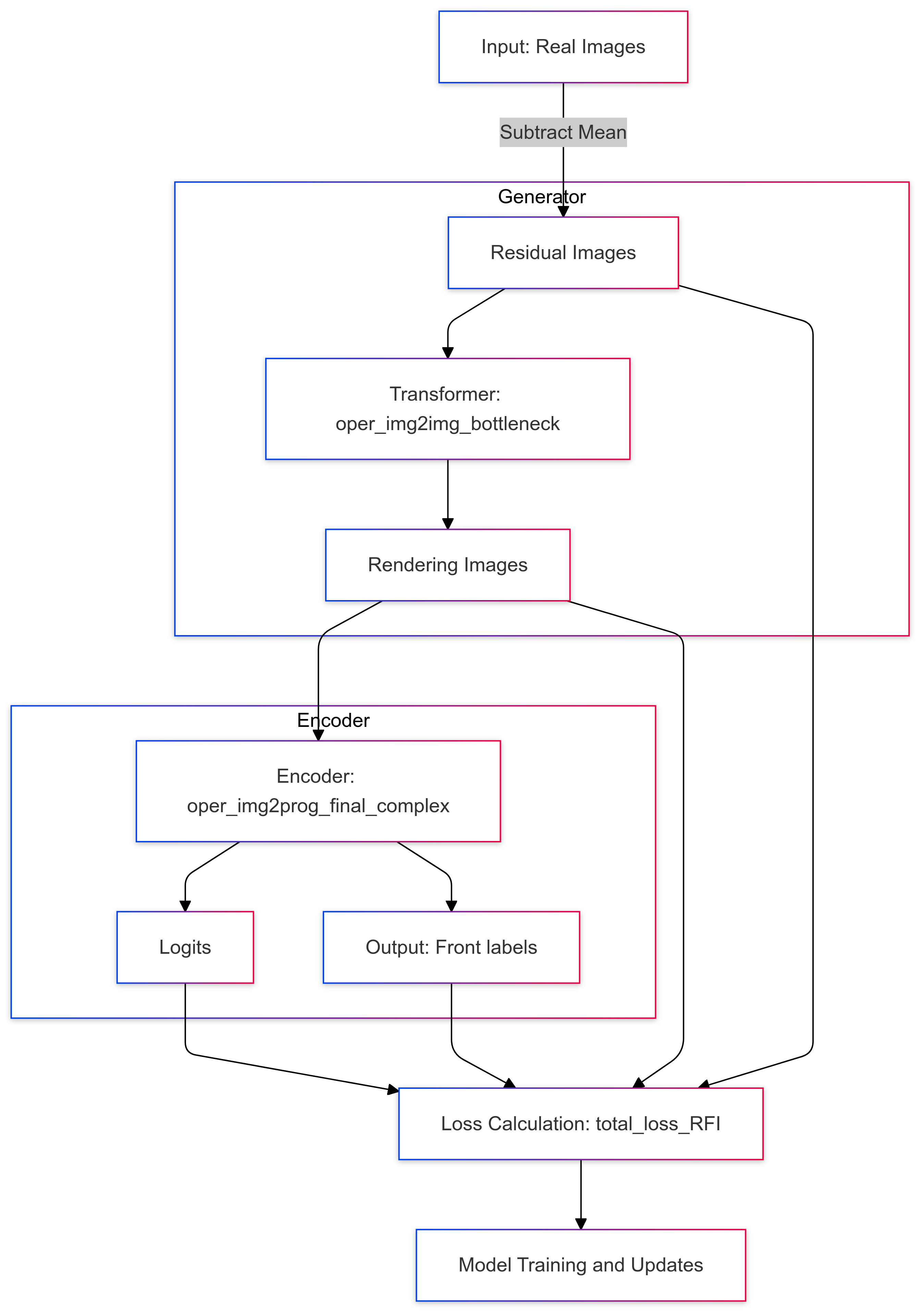}\label{fig:sub1}}
\hspace{20pt}
\subfloat[\centering]{\includegraphics[scale=0.5]{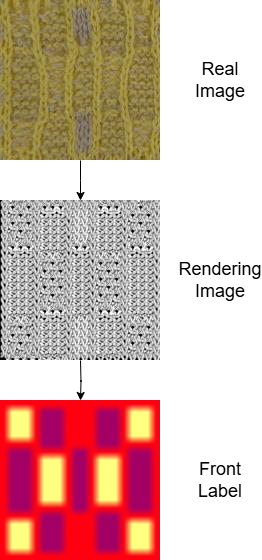}\label{fig:sub2}}
\end{adjustwidth}
\caption{Refiner+Img2prog architecture: (\textbf{a}) refiner+Img2prog diagram; (\textbf{b}) input and output.\label{FIG: refiner img2prog}}
\end{figure}

The \textit{generation phase} of our model processes the input \textit{real images} into \textit{front labels} in \mbox{two steps}: a \textit{refiner} is used to produce intermediate \textit{rendering images} and an Img2prog module to map these images to \textit{front labels} (Figure~\ref{FIG: refiner img2prog}). While the refiner acts as the \textit{generator}, and Img2prog as the \textit{encoder}, their specific roles are enhanced with innovations that improve the accuracy and robustness of the label prediction pipeline. This approach builds upon and improves Kaspar et al.'s work \cite{kaspar2019neural}.

This study improves upon Kaspar et al. \cite{kaspar2019neural} in two key areas. First, the precise mapping from \textit{real images} to \textit{rendering images} replaces Kaspar's indirect transfer approach, which aligns \textit{real images} to the rendering domain without creating specific one-to-one mappings. In contrast, our refiner directly maps each \textit{real image} to a corresponding predefined \textit{rendering image}, reducing ambiguity and improving the accuracy of downstream label predictions.

Second, this work introduces a stricter cross-entropy loss by eliminating the multiple-instance learning (MIL) tolerance used by Kaspar et al., which allowed neighboring labels to be considered correct. While MIL provided flexibility, it compromised precision, especially on imbalanced datasets where labels like \textit{FK} dominate (accounting for 75\% of data points)
. By enforcing stricter accuracy, our approach avoids overfitting to dominant labels and ensures improved performance across all classes \cite{kaspar2019neural}.

The refiner is a GAN model introduced by Shrivastava et al. (2017) \cite{shrivastava2017learning}. Their work proposed the idea of refining synthetic images into realistic ones using adversarial training. This foundational concept inspired the design of the \textit{refiner} used in our pipeline, which adapts their approach to refine residual images into accurate \textit{rendering images}.

The \textit{refiner}, implemented as \textit{oper\_img2img\_bottleneck}, refines \textit{residual images} derived from \textit{real images} into \textit{rendering images}, which serve as intermediary representations. Acting as the \textit{generator} in the pipeline, the \textit{refiner} operates through a series of distinct stages. Initially, during input preprocessing, a \textit{residual} image is generated by subtracting a mean value from the \textit{real image}, which normalizes the input and highlights key patterns for further processing. The encoder process then extracts features from the \textit{residual image} using convolutional layers and a stride of two to progressively downsample and encode essential information, aided by ReLU activations to ensure non-linearity. At the model's core, the bottleneck comprises residual blocks that preserve spatial dimensions while capturing deeper and more intricate features, crucial for generating accurate \textit{rendering images}. Finally, the decoder process reconstructs the processed features into a refined \textit{rendering image} using upsampling techniques, such as bilinear interpolation followed by convolutional layers. The output is a \textit{rendering image} that closely matches the ground truth, offering clear visual cues for subsequent tasks like label prediction.

The Img2prog module, implemented as \textit{oper\_img2prog\_final\_complex}, transforms the refined \textit{rendering images} output from the \textit{refiner} into \textit{front labels} that encode knitting instructions. Acting as the encoder in the pipeline, Img2prog begins with feature extraction, where convolutional layers reduce the spatial dimensions of the input while capturing its essential features. During this stage, intermediate outputs are preserved for use in skip connections. These skip connections integrate intermediate features, by transforming them through space-to-depth operations, reducing them via convolution, and then concatenating them to form a combined feature map. This approach ensures the fusion of both low-level and high-level information for enhanced feature learning. The residual blocks, similar to those in the refiner, further refine the combined features by processing the local and global dependencies within the knitting patterns. Finally, the output layer uses convolutional layers to map the refined features into \textit{logits} corresponding to the 14 \textit{front labels}. The predicted labels are derived through an \textit{argmax} operation, which determines the most probable class for each label.

By refining specific \textit{rendering images} with the \textit{refiner} and mapping them to \textit{front labels} with Img2prog, our method achieves significant improvements in precision and robustness. These two steps form the backbone of our \textit{generation phase}, addressing key shortcomings in Kaspar et al.'s approach and laying the foundation for subsequent \textit{inference-phase} processing.

Our project employs multiple loss functions to guide the training process effectively. These losses include cross-entropy, adversarial, perceptual, and style losses. Each loss component contributes uniquely to the generation of accurate and realistic knitting instructions and images.

Cross-entropy loss is a standard metric used in classification tasks to quantify the difference between the predicted probability distribution and the ground truth labels. The formula for cross-entropy loss is\vspace{-3pt}
\begin{linenomath}
\begin{equation}
\mathcal{L}_{\text{CE}} = - \sum_{i=1}^N y_i \log (\hat{y}_i)
\label{eq: cross entropy}
\end{equation}
\end{linenomath}
where \(y_i\) represents the ground truth label for class \(i\), \(\hat{y_i}\) is the predicted probability for class \(i\), and \(N\) denotes the total number of classes.

In this project, cross-entropy loss is applied to the \textit{logits} generated by the \textit{Img2prog} module to compare the predicted \textit{front labels} with the ground truth instructions. This loss is computed for both real and synthetic data. Unlike Kaspar et al. \cite{kaspar2019neural}, who implemented a multiple-instance learning (MIL) strategy to allow label tolerance, our approach strictly evaluates exact matches between predictions and the ground truth. This stricter evaluation avoids inflating accuracy metrics and ensures precise predictions, particularly for imbalanced datasets. Overall, cross-entropy loss is crucial for improving the model's ability to generate accurate knitting instructions as it effectively aligns the model's predictions with the target labels.

Adversarial loss, inspired by Generative Adversarial Networks (GANs) \cite{goodfellow2014generative}, is employed to enhance the realism of the generated \textit{rendering images}. We utilize a Conditional GAN (cGAN) framework \cite{mirza2014conditional}, where the discriminator distinguishes between real and generated images based on the input instructions.

The adversarial loss of a generator is computed using the least squares GAN approach \cite{mao2017least}:
\begin{linenomath}
\begin{equation}
\mathcal{L}_{\text{adv}} = \mathbb{E}_{x}[(D(G(x), y) - 1)^2]
\label{eq: adversal loss}
\end{equation}
\end{linenomath}
where \(x\) represents the input image, \(y\) is the conditioning information (instructions), \(G(x)\) is the image generated by the refiner, and \(D\) is the discriminator.

In our method, the \textit{refiner} serves as the generator, producing \textit{rendering images} from \textit{real images}, while the discriminator evaluates their authenticity. Specifically, the discriminator compares ground truth rendering images, or positive samples, to the generated \textit{rendering images} or ground truth \textit{real images}, which are considered negative samples. 
Adversarial loss forces the \textit{refiner} to generate \textit{rendering images} that are indistinguishable from real ones, ensuring they achieve a higher degree of visual realism. This improved realism enhances the quality of the generated outputs, making them more suitable for subsequent tasks such as accurate \textit{front label} prediction.

Perceptual loss evaluates the similarity between the generated and ground truth \textit{rendering images} in the feature space of a pre-trained VGG network \cite{simonyan2015very}. This loss builds upon the work of Gatys et al. (2016) \cite{gatys2016image} {on} neural style transfers. Perceptual loss, specifically when focusing on style similarity, is defined as follows:\vspace{-3pt}
\begin{linenomath}
\begin{equation}
\mathcal{L}_{\text{style}} = \sum_{l} w_l \| G^l(\hat{x}) - G^l(x) \|_F^2
\label{eq: style loss}
\end{equation}
\end{linenomath}
where \(G^l(\cdot)\) represents the Gram matrix of the feature maps from layer \(l\), \(\hat{x}\) and \(x\) are the generated and ground truth images, \(w_l\) denotes the weight assigned to each layer, and \(\|\cdot\|_F\) is the Frobenius norm.

In our method, perceptual loss evaluates the features extracted from layers of the VGG network. By comparing these features, the loss ensures that the generated \textit{rendering images} maintain their structural and textural consistency with the ground truth. This encourages the model to produce outputs that not only visually resemble the target images but also retain the stylistic and structural details necessary for accurate instruction generation.

Syntax loss is designed to enforce syntactic correctness in the predicted knitting instructions. It penalizes invalid transitions between stitches based on a predefined syntax of knitting patterns. This loss function does not have a straightforward mathematical formula but operates by referencing a transition matrix that defines valid stitch combinations.

In our method, the predicted instruction logits are compared against these allowable transitions to ensure logical and structural compliance with knitting patterns. This approach guarantees that the generated instructions are not only accurate at the individual stitch level but also syntactically valid. By adhering to these constraints, the syntax loss ensures that the resulting instructions can be feasibly manufactured, making them both practical and consistent with real-world knitting requirements.

The total loss function for the \textit{generator phase} combines several components to ensure the generated \textit{front labels} are accurate, realistic, and syntactically valid. The total loss can be expressed as follows:
\begin{linenomath}
\begin{equation}
\mathcal{L}_{\text{total}} = \mathcal{L}_{\text{CE}} + \mathcal{L}_{\text{adv}} + \mathcal{L}_{\text{style}} + \mathcal{L}_{\text{syntax}}
\label{eq: total loss}
\end{equation}
\end{linenomath}
These components work together to achieve high-quality \textit{front label} predictions that not only align with the knitting patterns but are also realistic and compliant with \mbox{knitting syntax}.

\subsection{Complete Label Inference: Residual Model}
The \textit{inference phase} focuses on transforming the {20 $\times$ 20} 
 \textit{front label}, generated by the \textit{generator} (\textit{refiner+Img2prog}), into a {20 $\times$ 20} \textit{complete label}. This step is essential for producing knittable instructions, as the \textit{front label} only represents the visible front side of the fabric, omitting the back layers necessary to create comprehensive knitting instructions.

\begin{figure}[H]
\centering
\begin{adjustwidth}{-\extralength}{0cm}
\subfloat[\centering]{\includegraphics[scale=0.42]{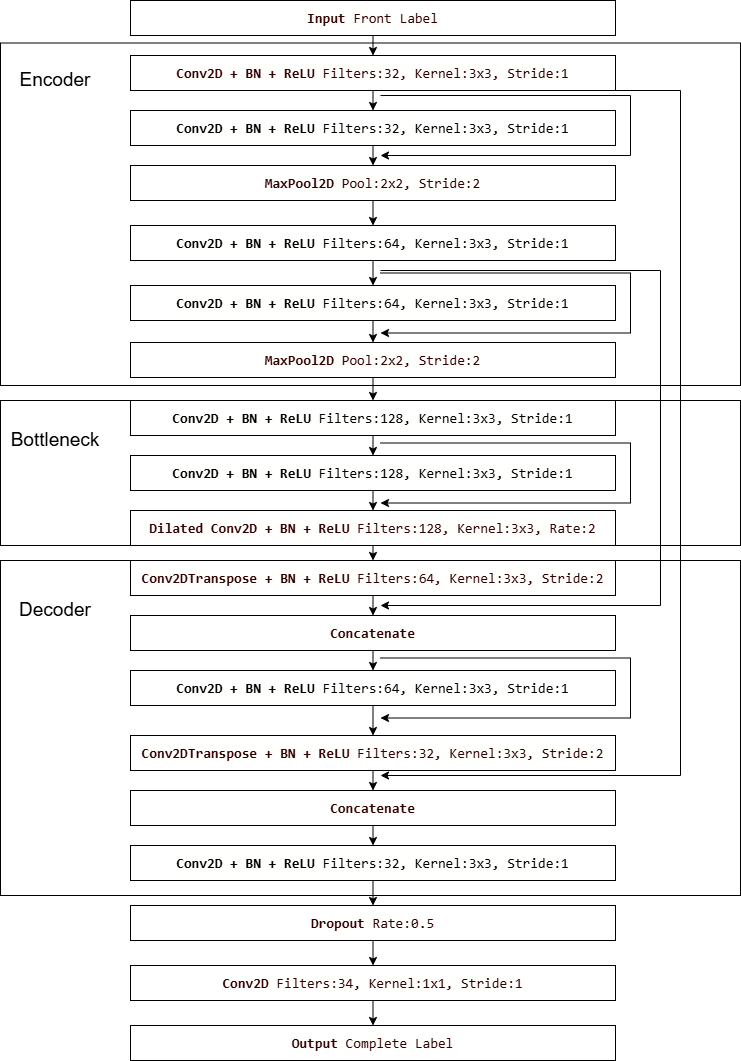}}
\hspace{20pt}
\subfloat[\centering]{\includegraphics[scale=0.55]{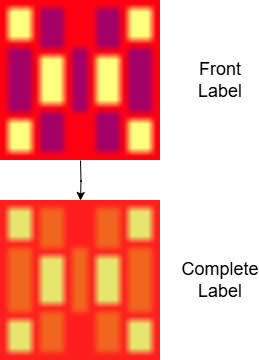}}
\end{adjustwidth}
\caption{Residual model architecture: (\textbf{a}) residual diagram; (\textbf{b}) input and output.\label{FIG: residual}}
\end{figure}

The \textit{residual model} transforms the \textit{front label} into a \textit{complete label }using a three-stage architecture that includes an \textit{encoder}, \textit{bottleneck}, and \textit{decoder} (Figure~\ref{FIG: residual}).

The \textit{encoder} begins by extracting features from the input using an initial convolution layer with Conv2D, BatchNorm, and ReLU activations, ensuring efficient feature representation. Residual blocks are then applied to process these features while preserving their spatial consistency, followed by max pooling to reduce their spatial resolution. Intermediate outputs are retained through skip connections, enabling the decoder to access earlier feature maps for precise reconstruction. The \textit{bottleneck} stage refines these features further using a deep residual block and includes dilated convolutions to expand the receptive field without increasing the number of parameters, capturing the long-range dependencies critical for accurate predictions. Finally, the \textit{decoder} reconstructs the \textit{complete label} by gradually upsampling the features to restore spatial resolution. It combines the earlier encoder outputs via skip connections to ensure that both high-level and low-level features contribute to the final output. A final convolution layer produces the 20 $\times$ 20 \textit{complete label}, encoding comprehensive knitting instructions with structural consistency.

The \textit{residual model} leverages residual blocks, skip connections, and dilated convolutions to ensure precise and efficient transformation of the \textit{front label} into a fully knittable \mbox{\textit{complete label}}.

The \textit{front label} exclusively contains information about the front-facing stitches, leaving out the structural details of the back side. A \textit{complete label} combines the information from both the front and back stitches, ensuring that the generated instructions are machine-compatible and create knittable outcomes. As demonstrated by Yuksel et al. (2012), in traditional stitch patterning, the label of the current stitch is strongly correlated with its four neighboring stitches (up, down, left, and right) \cite{yuksel2012stitch}. This intrinsic relationship validates our choice to incorporate \mbox{3 $\times$ 3 kernels} in the network architecture, as they effectively capture local spatial dependencies.

The \textit{residual model} utilizes residual blocks inspired by He et al.'s ResNet architecture, which addresses the vanishing gradient problem in deep networks \cite{he2016deep}. Each block consists of convolutional layers with skip connections, enabling gradients to flow directly through the network and allowing the model to learn more effectively. This architecture is particularly well suited to tasks requiring spatial consistency, such as inferring \textit{complete labels} from \textit{front labels}, where minor errors can propagate throughout the knitting instruction set.

To train the \textit{inference phase}, we employed a cross-entropy loss function (Equation (\ref{eq: cross entropy})). Unlike the \textit{generator phase}, where pixel-level differences in \textit{real images} and \textit{front labels} have a minimal impact on accuracy, the \textit{inference phase} demands precise stitch-level predictions. Introducing multiple-instance learning (MIL) techniques, as done by\linebreak Kaspar et al. (2019) \cite{kaspar2019neural}, would exacerbate the prediction error by neglecting the fine-grained differences between adjacent stitches. Given the small input resolution (a 20 $\times$ 20 \textit{front label}), such approximations lead to substantial inaccuracies in the \textit{complete label}. By strictly adhering to a standard cross-entropy loss, our model effectively balances the interdependencies between adjacent stitches and enforces stricter accuracy requirements.

\section{Experimental Setup and Results}\label{sec4}
Our model supports four key usage scenarios, as detailed in Appendix \ref{chap:usage_scenarios}. The first scenario focuses on \textit{front label} generation, where real knitting images are processed using the \textit{refiner} and Img2prog modules to produce \textit{front labels}. The second scenario, \textit{complete label generation (unknown yarn type)}, generates a \textit{complete label} without prior knowledge of the input yarn type by utilizing the full pipeline, including inference. The third scenario, \textit{complete label generation (known yarn type)}, further optimizes this process by leveraging yarn-specific residual models, distinguishing between the single-yarn (\textit{sj}) and multi-yarn (\textit{mj}) categories. Finally, the fourth scenario, \textit{complete label generation (using ground truth front label)}, directly uses ground truth \textit{front labels} as input to produce \textit{complete labels}, reducing the model's dependence on the \textit{front label generation} step. Each scenario includes detailed commands for execution and training, with TensorBoard support for monitoring the model's performance and visualizing its results. For more information, see Appendix \ref{chap:usage_scenarios}.

The experiments carried out for this project were conducted on a system configured with an NVIDIA RTX 2070 GPU running on Windows 11 with WSL2. The Ubuntu 18.04.6 operating system was installed within WSL2 to set up the necessary Linux-based environment. The deep learning framework used for this project was TensorFlow 1.11, with Python 3.6, CUDA Toolkit 9.0, and cuDNN 7.1 used for GPU acceleration. 
The following sections detail the environment configuration process and essential software installations. The full setup process, with detailed and reproducible commands, is documented in Appendix \ref{chap:environment_configuration}.


To evaluate the model's performance comprehensively, we designed four scenarios, each reflecting varying levels of input information and output requirements. The first scenario focuses on generating \textit{front labels} from \textit{real images} using the \textit{generation phase }(\textit{refiner} + Img2prog), which forms the basis for subsequent tasks. The second scenario generates \textit{complete labels} without prior knowledge of yarn types by first producing the \textit{front label} and then inferring the \textit{complete label}. In the third scenario, in which yarn type information (\textit{sj} or \textit{mj}) is known, the process uses a yarn-specific residual model for \textit{complete label} generation. Finally, the fourth scenario bypasses \textit{front label} generation entirely by directly inputting a ground truth \textit{front label} into the residual model to predict the \textit{complete label}. 

These scenarios provide a structured framework for analyzing the model's adaptability and performance across a diverse range of tasks.

\subsection{Generation Phase Evaluation---Scenario 1}

Table~\ref{tab:models of generation phase} summarizes the results from different models used in the \textit{generation phase}. The metrics evaluated include \textit{sample size}, \textit{parameter count} (indicating model complexity), \textit{training time} (in hours), and \textit{F1-score}, which acts as the performance indicator. The \textit{F1-score} was chosen due to the highly imbalanced label distribution in the dataset, ensuring a robust evaluation of the model's performance across all classes.


\begin{table}[H] 
\caption{Results of models used for generation phase.\label{tab:models of generation phase}}
\begin{adjustwidth}{-\extralength}{0cm}
\setlength{\cellWidtha}{\fulllength/5-2\tabcolsep+0.7in}
\setlength{\cellWidthb}{\fulllength/5-2\tabcolsep-0.1in}
\setlength{\cellWidthc}{\fulllength/5-2\tabcolsep-0.1in}
\setlength{\cellWidthd}{\fulllength/5-2\tabcolsep-0.3in}
\setlength{\cellWidthe}{\fulllength/5-2\tabcolsep-0.2in}
\scalebox{1}[1]{\begin{tabularx}{\fulllength}{>{\raggedright\arraybackslash}m{\cellWidtha}>{\raggedright\arraybackslash}m{\cellWidthb}>{\raggedright\arraybackslash}m{\cellWidthc}>{\raggedright\arraybackslash}m{\cellWidthd}>{\raggedright\arraybackslash}m{\cellWidthe}}
\toprule
\textbf{Model} & \textbf{Sample Size} & \textbf{Params Count} * & \textbf{Time (h)} & \textbf{F1-Score} \\ 
\midrule
RFI\_complex\_a0.5 & 12,392 & 2,934,605 & 6.50 & 90.2\% \\
RFINet\_notran\_noaug\_newinst & 12,392 & 2,934,398 & 5.00 & 97.3\% \\
RFINet\_front\_xferln\_MIL\_160k & 4950 & 2,934,398 & 3.00 & 82.1\% \\
RFINet\_front\_xferln\_160k ** & {4950} & {2,934,398} & {3.00} & {83.1\%} \\
\bottomrule
\end{tabularx}}
\end{adjustwidth}
\noindent{\footnotesize{{*} Params Count indicates the model's complexity. {**} This is the selected model.}}

\end{table}

The first two models, \textit{RFI\_complex\_a0.5} and \textit{RFI\_notran\_noaug\_newinst}, were trained on Kaspar's dataset of 12,392 single-yarn (\textit{sj}) samples. The baseline model, \textit{RFI\_complex\_a0.5}, used a label space with 17 classes, which made visual differentiation challenging and reduced its flexibility in generating extended knittable instructions. In contrast, \textit{RFI\_notran\_noaug\_newinst} addressed these issues by reducing the label space to 14 \textit{front labels}, simplifying the classification task and improving its adaptability for \textit{complete label} generation. Additionally, it used paired input--output data, explicitly mapping \textit{real images} to \textit{rendering images}, and introduced \textit{transfer images} generated with m1plus software as augmented data to compensate for the scarcity of real images. These improvements increased the \textit{F1-score} from 90.2\% to 97.3\% while reducing the model's complexity and training time.

The next two models, \textit{RFINet\_front\_xferln\_MIL\_160k} and \textit{RFINet\_front\_xferln\_160k}, were trained on a curated dataset of 4950 samples, consisting of 3000 \textit{sj} and 1950 \textit{mj} samples. Both models leveraged transfer learning, using pretrained weights from \textit{RFI\_notran\_noaug\_newinst} to enable faster convergence. The \textit{RFINet\_front\_xferln\_MIL\_160k} model incorporated MIL into its cross-entropy loss, allowing tolerance for neighboring labels. While this approach improved its fault tolerance, it reduced precision due to over-generalization, achieving an \textit{F1-score} of 82.1\%. By eliminating MIL and enforcing stricter label predictions, the \textit{RFINet\_front\_xferln\_160k} model achieved a higher \textit{F1-score} of 83.1\%, particularly improving its predictions for underrepresented labels.

We selected \textit{RFINet\_front\_xferln\_160k} as the \textit{generation phase} model for Scenario 1 (\textit{front label} prediction) based on three factors. First, it was trained on a balanced dataset of 4950 samples, ensuring compatibility with multi-yarn (\textit{mj}) data, which were limited in number. Second, its stricter cross-entropy loss resulted in improved accuracy when handling the imbalanced label distribution. Finally, the use of 14 \textit{front labels} provided the flexibility needed for generating \textit{complete labels}, making this model a robust and versatile solution for downstream tasks.

The performance of the \textit{RFINet\_front\_xferln\_160k} model in predicting each stitch in Scenario 1 (\textit{sj} + \textit{mj}) is presented in Table \ref{tab:results_generation_phase}. The results demonstrate the model's success in identifying frequently occurring labels, 
such as \textit{FK} and \textit{BK}, for which it achieved \textit{F1-score}s of 90.5\% and 78.2\%, respectively. These high scores can be attributed to the dominance of these stitches in the dataset, allowing the model to make robust predictions.

\begin{table}[H] 
\caption{Results from generation phase for each stitch.\label{tab:results_generation_phase}}
\setlength{\cellWidtha}{\textwidth/3-2\tabcolsep-0in}
\setlength{\cellWidthb}{\textwidth/3-2\tabcolsep-0in}
\setlength{\cellWidthc}{\textwidth/3-2\tabcolsep-0in}
\scalebox{1}[1]{\begin{tabularx}{\textwidth}{>{\raggedright\arraybackslash}m{\cellWidtha}>{\raggedright\arraybackslash}m{\cellWidthb}>{\centering\arraybackslash}m{\cellWidthc}}
\toprule
\multirow{2.5}{*}{\textbf{Stitch} *} & \multicolumn{2}{c}{\textbf{Scenario 1 (\textit{sj} + \textit{mj})}} \\ 
\cmidrule(r){2-3}
 & \textbf{Count} & \textbf{F1-Score} \\ 
\midrule
FK & 1,484,133 & 90.5\% \\ 
BK & 209,577 & 78.2\% \\ 
T  & 87,510  & 53.6\% \\ 
H  & 41,059  & 67.8\% \\ 
M  & 37,223  & 35.5\% \\ 
E  & 166     & 0.0\%  \\ 
V  & 1471   & 34.9\% \\ 
VR & 25,359  & 69.1\% \\ 
VL & 25,733  & 64.1\% \\ 
X(R) & 7031 & 68.5\% \\ 
X(L) & 7043 & 59.8\% \\ 
O  & 18,933  & 43.0\% \\ 
Y  & 22,904  & 63.3\% \\ 
FO & 11,858  & 30.8\% \\ 
\bottomrule
\end{tabularx}}
\noindent{\footnotesize{{*} Stitch labels and images can be observed in  Figure~\ref{FIG: maps}.}}
\end{table}

However, the analysis highlights significant challenges encountered with underrepresented labels. 
For instance, stitch E, for which there were only 166 samples, achieved an \textit{F1-score} of 0.0\%, showing the model's inability to predict this rare label. Similarly, stitch V, for which there were only 1,471 samples, achieved a low \textit{F1-score }of 34.9\%, indicating the adverse impact of insufficient training data. In contrast, labels such as VR, VL, and X(R), which had moderate sample counts, achieved \textit{F1-scores} of 69.1\%, 64.1\%, and 68.5\%, respectively, reflecting the model's ability to generalize when the distribution of data is relatively balanced.

These findings emphasize the importance of addressing data imbalances to improve predictions for less common labels. Increasing the sample size for stitches like E and V through targeted data augmentation or rebalancing efforts could enhance the model's performance on these stitches. 
Overall, the model shows strong results for common stitches and lays a promising foundation for further improvements in handling underrepresented labels.

\subsection{Inference Phase Evaluation---Scenarios 2, 3, and 4}

Table \ref{tab:models of inference phase} presents the results for various models used in the \textit{inference phase} across three scenarios: Scenario 2 (unknown yarn type), Scenario 3 (known yarn type), and Scenario 4 (ground truth \textit{front label}). The columns are similar to those in the \textit{generation phase} table, detailing sample size, parameter count, training time, and \textit{F1-score}. This table highlights several trends and key observations about the performance of the \textit{inference phase} models.

\begin{table}[H] 
\caption{Results for models used in inference phase.\label{tab:models of inference phase}}

\begin{adjustwidth}{-\extralength}{0cm}
\setlength{\cellWidtha}{\fulllength/5-2\tabcolsep+1.1in}
\setlength{\cellWidthb}{\fulllength/5-2\tabcolsep-0.1in}
\setlength{\cellWidthc}{\fulllength/5-2\tabcolsep-0.2in}
\setlength{\cellWidthd}{\fulllength/5-2\tabcolsep-0.4in}
\setlength{\cellWidthe}{\fulllength/5-2\tabcolsep-0.4in}
\scalebox{1}[1]{\begin{tabularx}{\fulllength}{>{\raggedright\arraybackslash}m{\cellWidtha}>{\centering\arraybackslash}m{\cellWidthb}>{\centering\arraybackslash}m{\cellWidthc}>{\centering\arraybackslash}m{\cellWidthd}>{\centering\arraybackslash}m{\cellWidthe}}
\toprule
\textbf{Model} & \textbf{Sample Size} & \textbf{Params Count} & \textbf{Time (h)} & \textbf{F1-Score} \\ 
\midrule
RFINet\_complete\_MIL & 4950 & 2,935,778 & 4.67 & 71.6\% \\
RFINet\_complete & 4950 & 2,935,778 & 4.67 & 80.8\% \\
xfer\_complete\_frompred\_2lyr\_MIL & 4950 & 21,026 & 2.75 & 39.4\% \\
xfer\_complete\_frompred\_2lyr & 4950 & 21,026 & 2.75 & 52.7\% \\
xfer\_complete\_frompred\_5lyr & 4950 & 1,585,422 & 3.00 & 78.1\% \\
xfer\_complete\_frompred\_residual * & {4950} 
 & 872,034 & 3.00 & 85.9\% \\
xfer\_complete\_frompred\_unet & 4950 & 279,138 & 3.00 & 83.9\% \\
\midrule
xfer\_complete\_frompred\_2lyr\_sj & 3000 & 21,026 & 1.75 & 95.0\% \\
xfer\_complete\_frompred\_residual\_sj * & 3000 & 872,034 & 1.75 & 97.0\% \\
xfer\_complete\_frompred\_unet\_sj & 3000 & 279,138 & 1.75 & 96.2\% \\
xfer\_complete\_frompred\_2lyr\_mj & 1950 & 21,026 & 1.00 & 74.0\% \\
xfer\_complete\_frompred\_residual\_mj * & 1950 & 872,034 & 1.00 & 90.2\% \\
xfer\_complete\_frompred\_unet\_mj & 1950 & 279,138 & 1.00 & 84.2\% \\
\midrule
xfer\_complete\_fromtrue\_2lyr\_sj & 3000 & 21,026 & 1.75 & 98.4\% \\
xfer\_complete\_fromtrue\_residual\_sj * & 3000 & 872,034 & 1.75 & 99.8\% \\
xfer\_complete\_fromtrue\_unet\_sj & 3000 & 279,138 & 1.75 & 99.3\% \\
xfer\_complete\_fromtrue\_2lyr\_mj & 1950 & 21,026 & 1.00 & 86.3\% \\
xfer\_complete\_fromtrue\_residual\_mj * & 1950 & 872,034 & 1.00 & 96.0\% \\
xfer\_complete\_fromtrue\_unet\_mj & 1950 & 279,138 & 1.00 & 95.4\% \\
\bottomrule
\end{tabularx}}
\end{adjustwidth}
\noindent{\footnotesize{{*} These are the selected models for each scenario.}}
\end{table}

\subsubsection{Scenario 2: Complete Label Generation (Unknown Yarn Type)}
In this scenario, where predicted \textit{front labels} are used as input without prior knowledge of the yarn type (\textit{sj} or \textit{mj}), two key observations emerge. First, the impact of MIL techniques is notable. Models incorporating MIL, such as \textit{RFINet\_complete\_MIL} and \textit{xfer\_complete\_frompred\_2lyr\_MIL}, underperform compared to their non-MIL counterparts. For instance, \textit{RFINet\_complete\_MIL} achieves an \textit{F1-score} of only 71.6\%, while the stricter \textit{RFINet\_complete} has an improved score of 80.8\%. Similarly, \textit{xfer\_complete\_frompred\_2lyr\_MIL} scores 39.4\%, whereas its non-MIL variant achieves 52.7\%. This discrepancy highlights the unsuitability of MIL for the \textit{inference phase}, where precise stitch-level predictions are essential, aligning with previous findings on MIL's limitations in handling fine-grained spatial dependencies at a 20 $\times$ 20 resolution.

Second, the effectiveness of the new CNN architectures becomes evident. Models using CNN-based architectures, such as the 2-layer, 5-layer, residual, and UNet variants, demonstrate varying levels of performance. Among these, the residual model (\textit{xfer\_complete\_frompred\_residual}) achieves the best balance between accuracy and complexity, with an \textit{F1-score} of 85.9\%. It outperforms \textit{RFINet\_complete} (80.8\%) and delivers comparable results to more complex models like \textit{xfer\_complete\_frompred\_5lyr} (78.1\%). These results emphasize the residual model’s efficiency and effectiveness in this task.

\subsubsection{Scenario 3: Complete Label Generation (Known Yarn Type)}
In this scenario, knowing the yarn type (\textit{sj} or \textit{mj}) enables the use of yarn-specific models, resulting in significant performance improvements. \textit{Residual} models, such as \textit{xfer\_complete\_frompred\_residual\_sj} and \textit{xfer\_complete\_frompred\_residual\_mj}, achieve \textit{F1-scores} of 97.0\% and 90.2\%, respectively, outperforming their Scenario 2 counterparts. This highlights the advantage of yarn-specific training in enhancing prediction accuracy. UNet models, including \textit{xfer\_complete\_frompred\_unet\_sj} and \textit{xfer\_complete\_frompred\_unet\_mj}, also perform well; however, the residual models maintain a slight edge in terms of accuracy, demonstrating their efficiency and robustness in this task.

\subsubsection{Scenario 4: Complete Label Generation (Using Ground Truth Front Label)}
In this ideal scenario, where the input is the ground truth \textit{front label}, bypassing the \textit{generation phase}, the \textit{inference phase} achieves its best theoretical performance. Residual models excel, with \textit{xfer\_complete\_fromtrue\_residual\_sj} achieving an \textit{F1-score }of 99.8\% and \textit{xfer\_complete\_fromtrue\_residual\_mj} reaching 96.0\%. These results validate the effectiveness of CNN-based approaches in capturing the spatial dependencies between neighboring stitches using 3 $\times$ 3 kernels, consistent with previous theoretical insights \cite{yuksel2012stitch}.

Key observations further highlight the pipeline's robustness. The incremental improvement from Scenario 2 to Scenario 4 demonstrates the \textit{inference phase’s} capacity to leverage additional input information, with its accuracy consistently increasing as more precise data are provided. Notably, despite the \textit{generation phase }achieving an \textit{F1-score} of 83.1\% in \textit{front label} prediction, the \textit{inference phase} compensates for these errors, achieving higher \textit{F1-scores} even in less favorable conditions, such as Scenario 2.

Moreover, the \textit{residual} model (\textit{xfer\_complete\_frompred\_residual}) consistently balances complexity and performance, outperforming 5-layer CNN models and achieving comparable results to UNet models with fewer parameters. This analysis underscores the \textit{inference phase’s} critical role in ensuring the pipeline’s overall accuracy and robustness by compensating for errors and leveraging spatial correlations to produce knittable \mbox{\textit{complete labels}}.


Table \ref{tab:results_inference_phase} presents the performance of the \textit{inference phase} models across Scenarios 2, 3, and 4, offering several insights. In Scenario 2, where the yarn type is unknown, common stitches like FK and BK perform well, with \textit{F1-scores} of 95.9\% and 82.6\%, respectively. However, rare stitches such as AO(2) (31.8\%) and FO(2) (32.3\%) struggle due to insufficient data. Additionally, multi-yarn-specific labels like V,HM achieve a modest accuracy (46.1\%), further underscoring the challenges posed by data scarcity.

\begin{table}[H] 
\small
\caption{Results from inference phase for each stitch.\label{tab:results_inference_phase}}
\begin{adjustwidth}{-\extralength}{0cm}
\setlength{\cellWidtha}{\fulllength/11-2\tabcolsep+0in}
\setlength{\cellWidthb}{\fulllength/11-2\tabcolsep-0in}
\setlength{\cellWidthc}{\fulllength/11-2\tabcolsep-0in}
\setlength{\cellWidthd}{\fulllength/11-2\tabcolsep-0in}
\setlength{\cellWidthe}{\fulllength/11-2\tabcolsep-0in}
\setlength{\cellWidthf}{\fulllength/11-2\tabcolsep-0in}
\setlength{\cellWidthg}{\fulllength/11-2\tabcolsep-0in}
\setlength{\cellWidthh}{\fulllength/11-2\tabcolsep-0in}
\setlength{\cellWidthi}{\fulllength/11-2\tabcolsep-0in}
\setlength{\cellWidthj}{\fulllength/11-2\tabcolsep-0in}
\setlength{\cellWidthk}{\fulllength/11-2\tabcolsep-0in}
\scalebox{1}[1]{\begin{tabularx}{\fulllength}{>{\raggedright\arraybackslash}m{\cellWidtha}>{\centering\arraybackslash}m{\cellWidthb}>{\centering\arraybackslash}m{\cellWidthc}>{\centering\arraybackslash}m{\cellWidthd}>{\centering\arraybackslash}m{\cellWidthe}>{\centering\arraybackslash}m{\cellWidthf}>{\centering\arraybackslash}m{\cellWidthg}>{\centering\arraybackslash}m{\cellWidthh}>{\centering\arraybackslash}m{\cellWidthi}>{\centering\arraybackslash}m{\cellWidthj}>{\centering\arraybackslash}m{\cellWidthk}}
\toprule
\multirow{2.5}{*}{\textbf{Stitch} *} & \multicolumn{2}{c}{\textbf{Scenario 2 (\emph{sj} + \emph{mj})}} & \multicolumn{2}{c}{\textbf{Scenario 3 (\emph{sj})}} & \multicolumn{2}{c}{\textbf{Scenario 3 (\emph{mj})}} & \multicolumn{2}{c}{\textbf{Scenario 4 (\emph{sj})}} & \multicolumn{2}{c}{\textbf{Scenario 4 (\emph{mj})}} \\ 
\cmidrule{2-11}
 & \textbf{Count} & \textbf{F1-Score} & \textbf{Count} & \textbf{F1-Score} & \textbf{Count} & \textbf{F1-Score} & \textbf{Count} & \textbf{F1-Score} & \textbf{Count} & \textbf{F1-Score} \\ 
\midrule
FK        & 920,228 & 95.9\% & 90,474 & 98.5\% & -       & 0.0\%  & 90,474 & 100.0\% & -      & 0.0\%  \\ 
BK        & 183,030 & 82.6\% & 13,375 & 95.2\% & 11,366  & 81.8\% & 13,375 & 99.7\%  & 11,366 & 92.8\% \\ 
T         & 87,698  & 65.2\% & 11,366 & 89.1\% & -       & 0.0\%  & 3207  & 98.9\%  & -      & 0.0\%  \\ 
H,M       & 15,618  & 66.2\% & -      & 0.0\%  & -       & 0.0\%  & 1001  & 97.1\%  & -      & 0.0\%  \\ 
M         & 22,342  & 65.0\% & 885    & 97.1\% & 2308   & 80.4\% & 885    & 97.1\%  & 2308  & 96.0\% \\ 
E,V(L)    & 15,901  & 64.5\% & 158    & 84.9\% & 158     & 94.8\% & 158    & 94.8\%  & -      & 0.0\%  \\ 
V,HM      & 1179   & 46.1\% & -      & 0.0\%  & -       & 0.0\%  & -      & 0.0\%   & -      & 0.0\%  \\ 
VR        & 8214   & 84.0\% & 809    & 92.1\% & 809     & 92.1\% & 809    & 99.7\%  & 809    & 99.7\% \\ 
VL        & 7919   & 82.0\% & 864    & 91.9\% & 864     & 91.9\% & 864    & 99.5\%  & 864    & 99.5\% \\ 
X(R)      & 7031   & 90.1\% & 741    & 95.9\% & 741     & 95.9\% & 741    & 99.6\%  & 741    & 99.6\% \\ 
X(L)      & 7043   & 90.0\% & 713    & 95.9\% & 713     & 95.9\% & 713    & 99.6\%  & 713    & 99.6\% \\ 
T(F)      & 25,024  & 78.5\% & 3207  & 89.1\% & -       & 0.0\%  & 3207  & 98.9\%  & -      & 0.0\%  \\ 
V,M       & 265     & 80.0\% & -      & 0.0\%  & 6       & 80.0\% & -      & 0.0\%   & -      & 0.0\%  \\ 
E,V(R)    & 15,703  & 85.0\% & 1747  & 89.1\% & -       & 0.0\%  & 1747  & 99.7\%  & -      & 0.0\%  \\ 
FK,MAK    & 536,224 & 88.9\% & -      & 0.0\%  & 107,069 & 94.6\% & -      & 0.0\%   & 107,069 & 99.0\% \\ 
FT,FMAK   & 65,413  & 88.9\% & -      & 0.0\%  & 13,106  & 92.1\% & -      & 0.0\%   & 13,106 & 92.6\% \\ 
Y,MATBK   & 22,904  & 63.3\% & 4741  & 98.9\% & -       & 0.0\%  & 4741  & 99.6\%  & -      & 0.0\%  \\ 
FO(2)     & 7324   & 32.3\% & -      & 0.0\%  & 1924   & 82.3\% & -      & 0.0\%   & 1,924  & 69.4\% \\ 
O(5),BK   & 7774   & 43.6\% & -      & 0.0\%  & 1500   & 51.9\% & -      & 0.0\%   & 1500  & 67.9\% \\ 
VR,FMAK   & 23,014  & 88.8\% & 4,419  & 95.9\% & -       & 0.0\%  & 4419  & 99.9\%  & -      & 0.0\%  \\ 
AO(2)     & 7767   & 31.8\% & -      & 0.0\%  & 1441   & 64.7\% & -      & 0.0\%   & 1441  & 64.0\% \\ 
\bottomrule
\end{tabularx}}
\end{adjustwidth}
\noindent{\footnotesize{{*} Stitch labels and images can be observed in  Figure~\ref{FIG: maps}.}}
\end{table}

In Scenario 3, where the yarn type is known, separating the single-yarn (\textit{sj}) and multi-yarn (\textit{mj}) datasets leads to significant performance improvements. For instance, FK leads to a 98.5\% accuracy on the \textit{sj} dataset, while FK,MAK reaches 94.6\% on the \textit{mj} dataset. The separation allows the model to specialize in stitches exclusive to either \textit{sj} or \textit{mj}, capturing their unique structural characteristics.

In the ideal case of Scenario 4, where ground truth \textit{front labels} are provided, the model achieves near-perfect results. \textit{FK} achieves 100\% accuracy with the \textit{sj} dataset, and \textit{FK,MAK }reaches 99.0\% accuracy with the \textit{mj} dataset, validating the theoretical feasibility of inferring \textit{complete labels} from accurate \mbox{\textit{front labels}}.

The key takeaways from this analysis include the benefits of separating yarn types during training, which significantly boosts stitch-specific accuracy, and the potential for near-ideal performance with perfect \textit{front label} inputs. Nonetheless, rare stitches remain challenging, highlighting the need for targeted data augmentation to improve the model's predictions of these stitches. The results also reinforce the importance of the structural insights gained from yarn type separation in enhancing the model’s specialization and performance.

\subsection{Case Study}
\begin{figure}[H]
\includegraphics[scale=0.4]{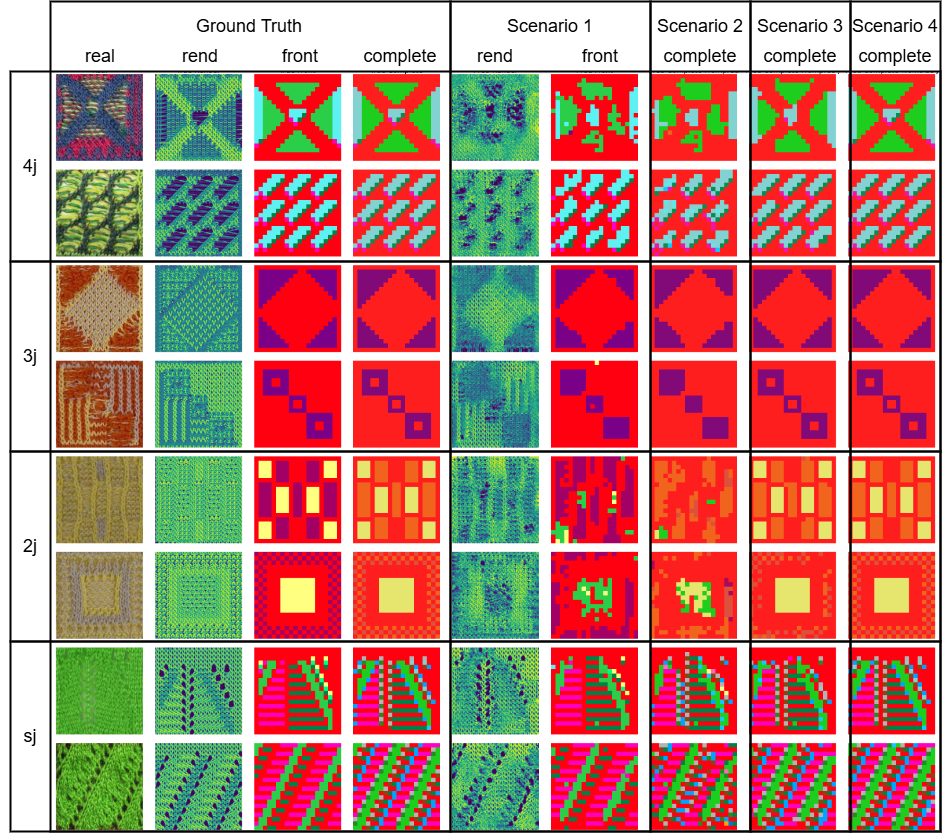}
\caption{Case study.\label{FIG: case study}}
\end{figure}

Figure \ref{FIG: case study} provides a detailed visualization of the results from multiple scenarios and yarn types, showcasing two samples each from the 4\emph{j}, 3\emph{j}, 2\emph{j}, and \emph{sj} categories. The figure is organized into columns that represent various stages: the ground truth (\textit{real image, rendering image, front label, and complete label}), predictions from Scenario 1 (\textit{rendering image }and \textit{front label}), predictions from Scenario 3 (\textit{complete label}, obtained using the predicted \textit{front label} and yarn-specific \textit{inference} models), and predictions from Scenario 4 (\textit{complete label} obtained using the ground truth \textit{front label}).

This analysis reveals several key observations. The model demonstrates a strong overall performance, accurately predicting \textit{front} and \textit{complete labels} across all yarn types. Notably, Scenario 3 highlights the model's ability to correct errors in \textit{front label} predictions, as seen in samples from the 4\emph{j} and 2\emph{j} yarn types. The model's predictions for \textit{sj} samples are near-perfect for both the front and complete labels, while \textit{mj} samples achieve high accuracy despite their more complex patterns. Finally, Scenario 4 serves as a benchmark, producing near-perfect \textit{complete labels}, underscoring the model's potential when given ideal inputs.

This case study highlights the system's robustness and adaptability, demonstrating its ability to handle diverse yarn types and patterns while effectively correcting errors, making it highly applicable to real-world knitting tasks.

\section{Discussion}\label{sec5}

The results demonstrate a clear progression in accuracy across the four scenarios, with Scenario 4 achieving near-perfect results while using ground truth \textit{front labels}. The \textit{generation phase}, powered by the \textit{RFINet\_front\_xferln\_160k} model, performs well on common stitches but struggles with rare ones, highlighting the importance of balanced datasets. The \textit{inference phase }compensates for \textit{generation phase} errors, showing a state-of-the-art performance with yarn-specific models and ground truth inputs.

Our case study reinforces these findings, showcasing the model's consistent performance across single-yarn (\textit{sj}) and multi-yarn (\textit{mj}) samples, with effective error correction in complex cases like 4\emph{j} and 2\emph{j} patterns. Key insights include the benefits of yarn type separation, strict loss functions, and residual CNN architectures, which enhance accuracy and robustness. However, data imbalance remains a challenge, particularly for rare stitches.

This research puts emphasis on the creation of structurally complete labels over colored complete labels, meaning the labels have true knittability. 
While the ultimate goal of reverse knitting includes full visual and structural fidelity, this research is limited to non-color-encoding structural outputs. Furthermore, challenges related to 3D garment shaping and material variability are beyond the scope of this study.

\section{Conclusions}\label{sec6}
This study addresses critical challenges in \textit{reverse knitting} by introducing a modular two-stage pipeline that separates \textit{front label }generation from \textit{complete label} inference. The pipeline leverages residual CNN models to capture spatial dependencies and generate precise knitting instructions, achieving \textit{F1-scores} of 83.1\% in Scenario 1 (\textit{generation phase}) and of up to 97.0\% in Scenario 3 (\textit{inference phase}). The \textit{inference phase's} ability to correct errors in the \textit{front label} predictions ensures the production of knittable instructions.

Future research should focus on expanding the relevant datasets, particularly those for multi-yarn samples; addressing label imbalance; 
and implementing data augmentation (e.g., rotation, brightness adjustments). Incorporating color recognition, flexible input--output dimensions, and advanced loss functions like focal loss could further enhance the system. Optimizing the pipeline for industrial knitting machines and extending it to 3D garment shaping and cross-domain textile processes (e.g., weaving, embroidery) will broaden its applicability and scalability. The current pipeline is constrained by fixed input dimensions (160 $\times$ 160 pixels) and stitch grids (20 $\times$ 20). Future research should explore handling variable input and output dimensions, leveraging object detection models like YOLO \cite{redmon2016you} to dynamically detect and label individual stitches, potentially supported by fine-grained stitch-level annotations. These advancements lay the foundation for fully automated, customizable textile manufacturing systems, meeting modern design and production demands.

\vspace{6pt} 

\authorcontributions{Conceptualization, X.Z. and H.S.; methodology, H.S. and M.L.; software, H.S.; validation, H.S., M.L., and S.C.; formal analysis, H.S.; investigation, X.Z. and M.L.; resources, X.Z.; data curation, S.C. and X.Z.; writing---original draft preparation, H.S.; writing---review and editing, M.L.; visualization, H.S.; supervision, M.L.; project administration, H.S. and M.L. All authors have read and agreed to the published version of the manuscript.}

\funding{This research received no external funding.}

\institutionalreview{Not applicable}

\informedconsent{Not applicable}

\dataavailability{The original data presented in the study are openly available on FigShare at \url{https://doi.org/10.6084/m9.figshare.28333379.v2}, {accessed on 18 February 2025}. The complete source code required for reproducing the results is available on GitHub at \url{https://github.com/SHolic/neural_inverse_knitting/tree/main}, {accessed on 18 February 2025}. Additionally, a demonstration video showcasing the inference procedure can be viewed on YouTube at \url{https://www.youtube.com/watch?v=GYR-Pck013s}, {accessed on 1 April 2025}.}

\acknowledgments{We would like to extend our heartfelt gratitude to Xingyu Zheng, Jiahui Shu, Tong Zhou, Ying Teng, and other summer interns from Shanghai Sanda University for their invaluable assistance throughout this project. Our deepest thanks also go to STOLL, created by KARL MAYER and SHIMASEIKI, 
	for generously providing equipment, technical support, and expertise that were critical to the success of this research.}

\conflictsofinterest{The authors declare no conflicts of interest.} 



\abbreviations{Abbreviations}{
The following abbreviations are used in this manuscript:\\

\noindent 
\begin{tabular}{@{}ll}
\emph{sj} & single-yarn\\
\emph{mj} & multi-yarn\\
MIL & multiple-instance learning\\
CNN & Convolutional Neural Network
\end{tabular}
}

\appendixtitles{no} 
\appendixstart
\appendix
\section[\appendixname~\thesection]{}\label{chap:usage_scenarios}
Our model supports four distinct use cases:
\begin{enumerate}
    \item {Scenario 1: Front Label Generation}
    \begin{itemize}
        \item Goal: Obtain a front label from a real image.
        \item Components Used: Generation phase (refiner + Img2prog).
        \item Command:
        \begin{verbatim}
            python main.py 
            --checkpoint_dir=./checkpoint/RFINet_front_xferln_160k 
            --training=False
        \end{verbatim}
        \vspace{-12pt}
        \item Input: Real image.
        \item Output: Front label.
        \item Training Command:
        \begin{verbatim}
            ./run.sh -g 0 
            -c ./checkpoint/RFINet_front_xferln_160k 
            --learning_rate 0.0005 
            --params discr_img=1,bvggloss=1,gen_passes=1,
            bloss_unsup=0,decay_steps=50000,
            decay_rate=0.3,bunet_test=3,
            use_tran=0,augment=0,bMILloss=0 
            --weights loss_D*=1.0,loss_G*=0.2 --max_iter 160000
        \end{verbatim}
         \vspace{-12pt}
        \item TensorBoard: We can use TensorBoard 
        to view various loss changes, generated images, multi-class confusion matrices, etc.
        \begin{verbatim}
            tensorboard 
            --logdir "./checkpoint/RFINet_front_xferln_160k/val"
        \end{verbatim}
    \end{itemize}
     \vspace{-12pt}
    \item Scenario 2: Complete Label Generation (Unknown Yarn Type)
    \begin{itemize}
        \item Goal: Obtain a complete label from a real image without prior knowledge of its \textit{sj/mj }classification.
        \item Components Used:
        \begin{itemize}
            \item Generation phase (refiner + Img2prog) for front label generation.
            \item Inference phase for complete label prediction.
        \end{itemize}
        \item Command:
        \begin{verbatim}
            python xfernet.py test 
            --checkpoint_dir ./checkpoint/xfer_complete_frompred
            _residual
            --model_type residual 
            --dataset default 
            --input_source frompred
        \end{verbatim}
         \vspace{-12pt}
        \item Input: Real image.
        \item Output: Complete label.
    \end{itemize}
    \item Scenario 3: Complete Label Generation (Known Yarn Type)
    \begin{itemize}
        \item Goal: Obtain a complete label with knowledge of the \textit{sj/mj} classification of the input image.
        \item Components Used:
        \begin{itemize}
            \item Generation phase (refiner + Img2prog).
            \item Yarn-specific residual model for inference phase. 
        \end{itemize}
        \item Command:
        \begin{verbatim}
            python xfernet.py test 
            --checkpoint_dir ./checkpoint/xfer_complete_frompred_sj 
            --model_type residual 
            --dataset sj 
            --input_source frompred
        \end{verbatim}
         \vspace{-12pt}
        Or for \textit{mj} data:
        \begin{verbatim}
            python xfernet.py test 
            --checkpoint_dir ./checkpoint/xfer_complete_frompred_mj 
            --model_type residual 
            --dataset mj 
            --input_source frompred
        \end{verbatim}
            \vspace{-12pt}
        \item Input: Real image.
        \item Output: Complete label.
    \end{itemize}
    \item Scenario 4: Complete Label Generation (Using Ground Truth Front Label)
    \begin{itemize}
        \item Goal: Generate complete labels using a ground truth front label and knowledge of yarn type.
        \item Components Used: Yarn-specific residual model.
        \item Command:
        \begin{verbatim}
            python xfernet.py test 
            --checkpoint_dir ./checkpoint/xfer_complete_fromtrue_sj 
            --model_type residual 
            --dataset sj 
            --input_source fromtrue
        \end{verbatim}
            \vspace{-12pt}
        Or for \textit{mj} data:
        \begin{verbatim}
            python xfernet.py test 
            --checkpoint_dir ./checkpoint/xfer_complete_fromtrue_mj 
            --model_type residual --dataset mj 
            --input_source fromtrue
        \end{verbatim}
            \vspace{-12pt}
        \item Input: Ground truth front label.
        \item Output: Complete label.
    \end{itemize}
\end{enumerate}

For Scenario 2 through to Scenario 4, if you wish to execute the training process, you simply need to change the parameter \textit{test} to \textit{train}. Additionally, you can utilize TensorBoard to monitor the loss progression and view the confusion matrix by specifying the given \textit{checkpoint\_dir}.

\section[\appendixname~\thesection]{}\label{chap:environment_configuration}
The experimental environment was set up using Miniconda for dependency management and package installation. The following steps outline the configuration process, with all commands provided for reproducibility:
\begin{enumerate}
    \item Install Miniconda
     \vspace{-3pt}
    \begin{verbatim}
        wget https://repo.anaconda.com/miniconda/Miniconda3-latest-Linux
        -x86_64.sh
        bash Miniconda3-latest-Linux-x86_64.sh
        source ~/.bashrc
    \end{verbatim}
         \vspace{-15pt}
    \item Create and Activate Python 3.6 Environment
    \begin{verbatim}
        conda create -n tf1.11 python=3.6 
        conda activate tf1.11
    \end{verbatim}
         \vspace{-15pt}
    \item Install GPU-Compatible TensorFlow and Its Dependencies. Install TensorFlow 1.11 and its associated dependencies, ensuring compatibility with the RTX 2070 and CUDA 9.0.
    \begin{verbatim}
        conda install tensorflow-gpu=1.11.0
        conda install numpy=1.15.3
        conda install scipy=1.1.0
    \end{verbatim}
         \vspace{-15pt}
    \item Install CUDA Toolkit and cuDNN
    \begin{verbatim}
        conda install cudatoolkit=9.0 cudnn=7.1
    \end{verbatim}
         \vspace{-15pt}
    \item Install Python Package Requirements. The required Python packages were installed using the \textit{requirements.txt} file provided in the project repository.
    \begin{verbatim}
        pip install -r requirements.txt
    \end{verbatim}
         \vspace{-15pt}
    \item Install ImageMagick for Image Processing. ImageMagick was used for image manipulation during the preprocessing stage. The following commands were used:
    \begin{verbatim}
        sudo apt update
        sudo apt install imagemagick
        sudo apt install zip unzip
    \end{verbatim}
         \vspace{-15pt}
    \item Set Up {Jupyter Notebook 6.4.3} 
 for Interactive Development. Jupyter Notebook was installed to facilitate interactive code testing and experimentation.
    \begin{verbatim}
        conda install jupyter
        jupyter notebook --ip=0.0.0.0 --no-browser
    \end{verbatim}
         \vspace{-15pt}
    \item Install Additional Libraries. For additional functionalities, the Scikit-learn library was installed.
    \begin{verbatim}
        conda install scikit-learn
    \end{verbatim}
\end{enumerate}

The experimental environment was designed to leverage the computational power of the RTX 2070 GPU and the stability of TensorFlow 1.11, ensuring compatibility with older dependencies and toolkits. The use of Miniconda allowed for efficient dependency resolutions, while WSL2 provided a seamless bridge between the Windows and Linux environments. By documenting the environmental setup with reproducible commands, this configuration can be easily replicated for future experiments or debugging purposes.

\begin{adjustwidth}{-\extralength}{0cm}

\reftitle{References}

\isAPAandChicago{}{%

}

\PublishersNote{}
\end{adjustwidth}
\end{document}